%% file: speech_parsing.tex
\documentclass[11pt,a4paper]{article}

\usepackage[hyperref]{naaclhlt2018}
\usepackage{graphicx}
\usepackage{times}
\usepackage{latexsym}
\usepackage{amsmath}
\usepackage{multirow}
\usepackage{enumitem}
\usepackage{url}
\usepackage{color}
\usepackage{textcomp}

\usepackage{microtype}

\setlist{nolistsep,leftmargin=*}

\newcommand{\ttcomment}[1]{\textcolor{red}{\bf \small [ #1 --TT]}}
\newcommand{\stcomment}[1]{\textcolor{blue}{\bf \small [ #1 --ST]}}
\newcommand{\mbcomment}[1]{\textcolor{cyan}{\bf \small [ #1 --MB]}}
\newcommand{\kgcomment}[1]{\textcolor{green}{\bf \small [ #1 --KG]}}
\newcommand{\klcomment}[1]{\textcolor{magenta}{\bf \small [ #1 --KL]}}
\newcommand{\klrm}[1]{\textcolor{magenta}{\bf \small [#1]}}
\newcommand{\mocomment}[1]{\textcolor{orange}{\bf \small [ #1 --MO]}}
\renewcommand{\ttcomment}[1]{}
\renewcommand{\stcomment}[1]{}
\renewcommand{\mbcomment}[1]{}
\renewcommand{\kgcomment}[1]{}
\renewcommand{\klcomment}[1]{}
\renewcommand{\mocomment}[1]{}

\renewcommand{\klrm}[1]{}

\newcommand\Mark[1]{\textsuperscript#1}

\def\vec#1{\ensuremath{\boldsymbol{{#1}}}}

\aclfinalcopy 
\def\aclpaperid{199} 
\title{Parsing Speech: A Neural Approach to Integrating \\ Lexical and Acoustic-Prosodic Information}
\date{}
\author{Trang Tran\thanks{\ \ Equal Contribution.}\ \Mark{1}, Shubham Toshniwal\footnotemark[1] \Mark{2}, 
Mohit Bansal\Mark{3},\\ 
{\bf Kevin Gimpel\Mark{2}}, 
{\bf Karen Livescu\Mark{2}},  
{\bf Mari Ostendorf\Mark{1}}\\[1em]
\Mark{1}Electrical Engineering, University of Washington\\
\Mark{2}Toyota Technological Institute at Chicago\\
\Mark{3}Department of Computer Science, UNC Chapel Hill\\[1em]
\texttt{\{ttmt001, ostendor\}@uw.edu, mbansal@cs.unc.edu},\\
\texttt{\{shtoshni, kgimpel, klivescu\}@ttic.edu}\\
}

\begin{document}
\maketitle

\begin{abstract}
In conversational speech, the acoustic signal provides cues that help listeners disambiguate difficult parses. For automatically parsing spoken utterances, we introduce a model that integrates transcribed text and acoustic-prosodic features using a convolutional neural network over energy and pitch trajectories coupled with an attention-based recurrent neural network that accepts text and prosodic features. We find that different types of acoustic-prosodic features are individually helpful, and together give statistically significant improvements in parse and disfluency detection F1 scores over a strong text-only baseline. For this study with known sentence boundaries, error analyses show that the main benefit of acoustic-prosodic features is in sentences with disfluencies, attachment decisions are most improved, and transcription errors obscure gains from prosody. 
\end{abstract}

\input{intro}
\input{model}
\input{experiment}
\input{result}

\input{analysis}
\input{related}

\input{conclusion}
\input{acknowledgement}


\bibliography{speech_parsing}
\bibliographystyle{acl_natbib}

\appendix
\input{appendix}

\end{document}

%% file: intro.tex
\section{Introduction}
\label{sec:intro}
While parsing has become a relatively mature technology for written text, parser performance on conversational speech lags behind\klrm{results on text}.  Speech poses challenges for parsing: \klrm{First,}transcripts may contain errors and lack punctuation;  
\klrm{Second,}even perfect transcripts can be difficult to handle because of disfluencies (restarts, repetitions, and self-corrections), filled pauses (``um'', ``uh''), interjections (``like''), parentheticals (``you know'', ``I mean''), and sentence fragments\klrm{that are common in spontaneous speech}. Some of these phenomena can be handled in standard grammars, but disfluencies typically require extensions of the model. Different approaches have been explored in both constituency parsing \cite{Charniak2001,Johnson2004} and dependency parsing \cite{Rasooli2013,Honnibal2014}. 

Despite these challenges, speech carries helpful extra information -- beyond the words  --  associated with the prosodic structure of an utterance and encoded via variation in timing and intonation. \klrm{Studies show that}Speakers pause in locations that are correlated with syntactic structure \cite{Grosjean79}, and listeners \klrm{are able to} use prosodic structure in resolving syntactic ambiguities \cite{Price1991}. Prosodic cues also signal disfluencies by marking the interruption point \cite{Shriberg94}. However, most speech parsing systems in practice take little advantage of these cues. 
Our study focuses on this last challenge, aiming to incorporate prosodic cues in a neural parser, handling disfluencies as constituents via a neural attention mechanism. 

A challenge of incorporating prosody in parsing is that multiple acoustic cues interact to signal prosodic structure, including pauses, lengthening, fundamental frequency modulation, and spectral shape. These cues also vary with the phonetic segment, emphasis, emotion and speaker, so feature extraction typically involves multiple\klrm{different} time windows and normalization techniques. The most successful constituent parsers have mapped these features to prosodic boundary posteriors by using labeled training data \cite{Kahn2005,Hale2006,Dreyer2007}. The approach proposed here takes advantage of advances in neural networks to automatically learn a good feature representation \klrm{for parsing} without the need to explicitly represent prosodic constituents. To narrow the scope of this work and facilitate error analysis, our experiments use known transcripts and sentence segmentation.

Our work offers the following contributions. We introduce a framework for directly integrating acoustic-prosodic features with text in a neural encoder-decoder parser that does not require hand-annotated prosodic structure. We demonstrate improvements in constituent parsing of conversational speech over a high-quality text-only parser
and provide analyses showing where prosodic features help and that assessment of their utility is affected by human transcription errors.

%% file: model.tex
\section{Task and Model Description}
\label{sec:models}

Our \klrm{encoder-attention-decoder} model maps a sequence of word-level input features to a linearized parse output sequence. 
The word-level input feature vector consists of the concatenation of (learnable) word embeddings $\vec{e_i}$ and several types of acoustic-prosodic features, described in Section \ref{ssec:features}.

\subsection{Task Setup}
We assume the availability of a training treebank of conversational speech (in our case, Switchboard-NXT \cite{Calhoun2010}) and corresponding constituent parses.  The transcriptions are preprocessed by removing punctuation and lower-casing all text to better mimic the speech recognition setting. 
Following Vinyals et al.\ \shortcite{Vinyals2015}, the parse trees are linearized, and pre-terminals are normalized as ``XX'' (see Appendix \ref{app:misc}).

\subsection{Encoder-Decoder Parser}
\begin{figure*}[t]
\centering
\centering
\includegraphics[trim={0 9.5cm 0 0},clip,scale=0.22]{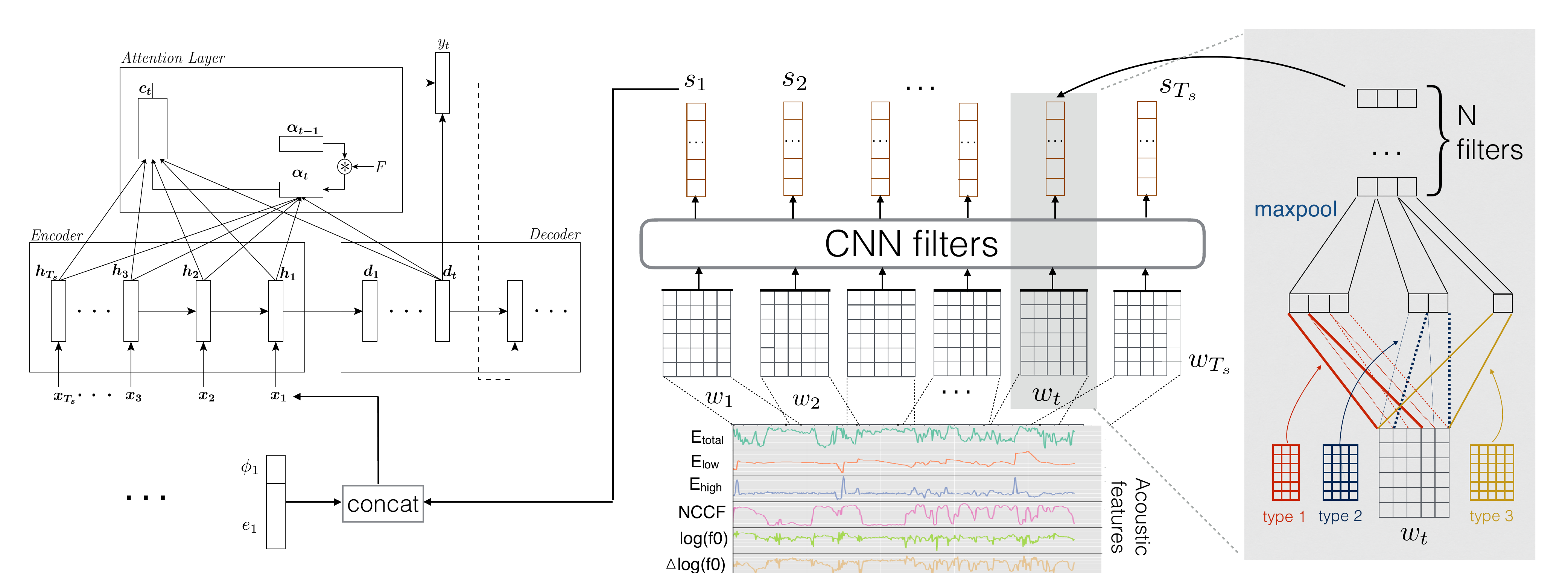}
\centering
\caption{Left -- An attention-based encoder-decoder reading the input $\vec{x}_{1}, \cdots, \vec{x}_{T_{s}}$, where $\vec{x}_i = [\vec{e}_i \; \vec{\phi}_i \; \vec{s}_i]$ is composed of word embeddings $\vec{e}_i$, prosodic features $\vec{\phi}_i$, and learned (CNN-based) features $\vec{s}_i$. The encoder reads the input in \emph{reverse} order and the decoder outputs the linearized parse $y_1, \cdots, y_t, \cdots$.
Right -- Detailed illustration of acoustic-prosodic feature learning module. CNN features are computed from input energy and pitch features; here the CNN filter parameters are $m=3$ and $w=[3,4,5]$.}
\label{fig:model-comb}
\end{figure*}

Our attention-based encoder-decoder model is similar to the one used by \newcite{Vinyals2015}. The \emph{encoder} is a deep long short-term memory recurrent neural network (LSTM-RNN)~\cite{Hochreiter1997} that reads in a word-level inputs,\footnote{As in \newcite{Vinyals2015} the input sequence is processed in reverse order, as shown in Figure~\ref{fig:model-comb}.} represented as a sequence of vectors $\vec{x} = (\vec{x}_1, \cdots, \vec{x}_{T_s})$, and outputs high-level features $\vec{h} = (\vec{h}_1, \cdots, \vec{h}_{T_s})$ where
$\vec{h}_{i}~=~\text{LSTM}(\vec{x}_{i}, \vec{h}_{i-1})$.\footnote{For brevity we omit the LSTM equations. The details can be found, e.g., in \newcite{Zaremba2014}.}

The \emph{parse decoder} is also a deep LSTM-RNN that predicts the linearized parse sequence $\vec{y} = (y_1, \cdots, y_{T_{o}})$ as follows:
$$P(\vec{y}|\vec{x}) = \prod_{t=1}^{T_o}P(y_{t}|\vec{h}, \vec{y_{< t}})$$
In attention-based models, the posterior distribution of the output $y_t$ at time step $t$ is given by:
$$P(y_{t}|\vec{h}, \vec{y_{< t}}) = \text{softmax}(\vec{W}_{s}[\vec{c}_{t}; \vec{d}_{t}] + \vec{b}_{s}),$$
where vector $\vec{b}_s$ and matrix $\vec{W}_{s}$ are learnable parameters; $\vec{c}_t$ is referred to as a \textit{context vector} that summarizes the encoder's output $\vec{h}$; and $\vec{d}_t$ is the decoder hidden state at time step $t$, which captures the previous output sequence context $\vec{y_{< t}}$.

\begin{align*}
u_{it} &= \vec{v}^\top \tanh(\vec{W}_1\vec{h}_i + \vec{W}_2\vec{d}_t + \vec{b}_a) \\
\vec{\alpha}_{t} & = \text{softmax}(\vec{u}_t) \qquad \vec{c}_t = \sum_{i=1}^{T_s} \alpha_{ti}\vec{h}_{i}
\end{align*}
where vectors $\vec{v}$, $\vec{b}_a$ and matrices $\vec{W}_1$, $\vec{W}_2$ are learnable parameters; $\vec{u}_t$ and $\vec{\alpha}_{t}$  are the attention score and attention weight vector, respectively, for decoder time step $t$. 

The above attention mechanism is only \emph{content}-based, i.e., it is only dependent on $\vec{h}_i$, $\vec{d}_t$. It is not \emph{location}-aware, i.e., it does not consider the ``location" of the previous attention vector. For parsing conversational text, location awareness is beneficial since disfluent structures can have duplicate words/phrases that may ``confuse'' the attention mechanism.

In order to make the model location-aware, the attention mechanism takes into account the previous attention weight vector $\vec{\alpha}_{t-1}$. In particular, we use the attention mechanism proposed by \newcite{Chorowski2015}, in which $\vec{\alpha}_{t-1}$ is represented via a feature vector 
$\vec{f}_t = \vec{F}\ast \vec{\alpha}_{t-1}$, 
where $\vec{F} \in \mathcal{R}^{k \times r}$ represents $k$ learnable convolution filters of width $r$. The filters are used for performing 1-$D$ convolution over $\vec{\alpha}_{t-1}$ to extract $k$ features $\vec{f}_{ti}$ for each time step $i$ of the input sequence. The extracted features are then incorporated in the alignment score calculation as: 
$$u_{it} = \vec{v}^\top \tanh(\vec{W}_1\vec{h}_i + \vec{W}_2\vec{d}_t + \vec{W}_f\vec{f}_{ti} + \vec{b}_a)$$
where $\vec{W}_{f}$ is another learnable parameter matrix.
Finally, the decoder \klrm{hidden }state $\vec{d}_t$ is computed as $\vec{d}_{t} = \text{LSTM}([\tilde{\vec{y}}_{t-1}; \vec{c}_{t-1}], \vec{d}_{t-1})$, where $\tilde{\vec{y}}_{t-1}$ is the embedding vector corresponding to the previous output symbol $y_{t-1}$. 
As we will see in Sec.~\ref{ssec:results-text}, the location-aware attention mechanism is especially useful for handling disfluencies.

\input{ap_features}

%% file: ap_features.tex
\subsection{Acoustic-Prosodic Features}
\label{ssec:features}

In previous work using encoder-decoder models for parsing~\cite{Vinyals2015,Luong2016}, vector
$\vec{x}_{i}$ is simply the word embedding $\vec{e}_i$ of the word at position $i$ of the input sentence. For parsing conversational speech, we can incorporate acoustic-prosodic features. Here we explore four types of features widely used in computational models of prosody: pauses, duration lengthening,  
fundamental frequency, and energy. Since prosodic cues are at sub- and multi-word time scales, 
they are integrated with the encoder-decoder using different mechanisms.

All features are extracted from transcriptions that are time-aligned at the word level.\footnote{The assumption of known word alignments is standard for prosodic feature extraction in many spoken language processing studies. Time alignments can be obtained as a by-product of recognition or from forced alignment.} We use time alignments associated with the corpus to be consistent with other studies.
In a small number of cases, the time alignment for a particular word boundary is missing. Some cases are due to tokenization. For example, contractions, such as \emph{don't} in the original transcript, are treated as separated words for the parser (\emph{do} and \emph{n't}), and the internal word boundary time is missing. In such cases, these internal times are estimated. In other cases, there are transcription mismatches that lead to missing time alignments, where we cannot estimate times. For the roughly 1\% of sentences where time alignments are missing, we simply back off to the text-based parser. 

\paragraph{Pause.} The pause feature vector $\vec{p}_i$ for word $i$ is the concatenation of pre-word pause feature $\vec{p}_{pre,i}$ and post-word pause feature $\vec{p}_{post,i}$, where each subvector is a learned embedding for 6 pause categories: no pause, missing, $0 < p \le 0.05$ s, $0.05 \; \textrm{s} < p \le 0.2$ s, $0.2 < p \le 1$ s, and $p > 1$ s (including turn boundaries). The bins are chosen based on the observed distribution (see Appendix \ref{app:misc}).
We did not use (real-valued) pause duration directly, for two main reasons: (1) to handle missing time alignments; and (2) duration of pause does not matter beyond a threshold (e.g. $p >$ 1 s).

\paragraph{Word duration.} Both word duration and word-final duration lengthening are strong cues to prosodic phrase boundaries \cite{Wightman+92,Pate2013}.
The word duration feature $\delta_i$ is
computed as the actual word duration divided by the mean duration of the word, clipped to a maximum value of 5. The sample mean is used for frequent words (count $\ge$ 15). For infrequent words we estimate the mean as the sum over the sample means for the phonemes in the word's dictionary pronunciation. We refer to the manually defined prosodic feature pair of $\vec{p}_i$ and $\delta_i$ as $\phi_i$.

\paragraph{Fundament frequency (f0) and Energy (E) contours (f0/E).}  
We use a CNN to automatically learn the mapping from the time series of f0/E features to a word-level vector. 
The contour features are extracted from 25-ms frames with 10-ms hops using Kaldi \cite{kaldi}.
Three f0 features are used: warped Normalized Cross Correlation Function (NCCF), log-pitch with Probability of Voicing (POV)-weighted mean subtraction over a 1.5-second window, and the estimated derivative (delta) of the raw log pitch. 
Three energy features are extracted from the Kaldi 40-mel-frequency filter bank features: $E_{total}$, the log of total energy normalized by dividing by the speaker side's max total energy; 
$E_{low}$, the log of total energy in the lower 20 mel-frequency bands, normalized by total energy, and $E_{high}$, the log of total energy in the higher 20 mel-frequency bands, normalized by total energy. Multi-band energy features are used as a simple mechanism  to capture articulatory strengthening at prosodic constituent onsets \cite{Fougeron97}.

Figure \ref{fig:model-comb} summarizes the feature learning approach. The f0 and E features are processed at the word level: each sequence of frames corresponding to a time-aligned word (and potentially its surrounding context) is convolved with $N$ filters of $m$ sizes (a total of $mN$ filters). The motivation for the multiple filter sizes is to enable the computation of features that capture information on different time scales. For each filter, we perform a 1-D convolution over the 6-dimensional f0/E features with a stride of 1. Each filter output is max-pooled, resulting in $mN$-dimensional speech features $\vec{s}_i$. Our overall acoustic-prosodic feature vector is the concatenation of $\vec{p}_i$, $\delta_i$, and $\vec{s}_i$ in various combinations. 

%% file: experiment.tex
\section{Experiments}
\label{sec:exp}
\subsection{Dataset} 
Our core corpus is Switchboard-NXT \cite{Calhoun2010}, a subset of the Switchboard corpus \cite{Godfrey1993}: 
2,400 telephone conversations between strangers; 
642 of these were hand-annotated with syntactic parses and further augmented with richer layers of annotation facilitated by the NITE XML toolkit \cite{Calhoun2010}. Our sentence segmentations and syntactic trees are based on the annotations from the Treebank set, with a few manual corrections from the NXT release. This core dataset consists of 100K sentences, totaling 830K tokens forming a vocabulary of 13.5K words.
We use the time alignments available from NXT, which is based on a corrected word transcript that occasionally differs from the Treebank, leading to some missing time alignments.
We follow the sentence boundaries defined by the parsed data available,\footnote{Note that these sentence units can be inconsistent with other layers of Switchboard annotations, such as \emph{slash units}.} and the data split (90\% train; 5\% dev; 5\% test) defined by related work done on Switchboard \cite{Charniak2001,Kahn2005,Honnibal2014}.

\subsection{Evaluation Metrics and Baselines}
\label{ssec:metrics}

The standard evaluation metric for constituent parsing is the \emph{parseval} metric which uses bracketing precision, recall, and F1, as in the canonical implementation of EVALB.\footnote{\url{http://nlp.cs.nyu.edu/evalb/}} 
For written text, punctuation is sometimes represented as part of the sequence and impacts the final score, 
but for speech the punctuation is not explicitly available so it does not contribute to the score. Another challenge of transcribed speech is the presence of disfluencies. Speech repairs are indicated under ``EDITED'' nodes in Switchboard parse trees, which include structure under these nodes that is not of interest for simple text clean-up. Therefore, some studies report
\emph{flattened-edit parseval} F1 scores (``flat-F1''), which is \emph{parseval} computed on trees where the structure under edit nodes has been eliminated so that all leaves are immediate children. We report both scores for the baseline text-only model showing that the differences are small, then use the standard \emph{parseval} F1 score for most results.\footnote{A variant of the ``flat-F1'' score is used in \cite{Charniak2001,Kahn2005}, which uses a relaxed edited node precision and recall but also ignores filled pauses.}

Disfluencies are particularly problematic for statistical parsers, as explained by \newcite{Charniak2001}, and some systems incorporate a separate disfluency detection stage. For this reason, and because it is useful for understanding system performance, most studies also report disfluency detection performance, which is measured in terms of the F1 score for detecting whether a word is in an edit region. Our approach does not involve a separate disfluency detection stage, but identifies disfluencies implicitly via the parse structure. 
Consequently, the disfluency detection results are not competitive with work that directly optimize for disfluency detection. We report disfluency detection scores primarily as a diagnostic.  

Most previous work on integrating prosody and parsing has used the Switchboard corpus, but it is still difficult to compare results because of differences in constraints, objectives and the use of constituent vs.\ dependency structure, as discussed further in Section \ref{sec:related}. The most relevant prior studies (on constituent parsing) that we compare to are a bit old. The text-only result from our neural parser represents a stronger baseline and is important for decoupling the impact of prosody vs.\ the parsing framework.

\subsection{Model Training and Inference}
\label{ssec:training}
Both the encoder and decoder are 3-layer deep LSTM-RNNs with 256 hidden units in each layer. For the location-aware attention, the convolution operation uses 5 filters of width 40 each. We use 512-dimensional embedding vectors to represent words and linearized parsing symbols, such as ``(S".\footnote{The number of layers, dimension of hidden units, dimension of embedding, and convolutional attention filter parameters of the text-only parser were explored in earlier experiments on the development set and then fixed as described.}

A number of configurations are explored for the acoustic-prosodic features, tuning based on dev set parsing performance. Pause embeddings are tuned over $\{4, 16, 32\}$ dimensions. For the CNN, we try different configurations of filter widths $w \in \{[10,25,50], [5,10,25,50]\}$ and number of filters $N \in \{16,32,64,128\}$ for each filter width.\footnote{Note that a filter of width 10 has size $6\times 10$, since the features are of dimension 6.} These filter size combinations are chosen to capture f0 and energy phenomena on various levels: $w=5, 10$ for sub-word, $w=25$ for word, and $w=50$ for word and extended context. Our best model uses 32-dimensional pause embeddings and $N=32$ filters of widths $w = [5,10,25,50]$, which corresponds to $m=4$ and 128 filters.

For optimization we use Adam~\cite{Kingma2014} with a minibatch size of 64. The initial learning rate is $0.001$ which is decayed by a factor of 0.9 whenever training loss, calculated after every 500 updates, degrades relative to the worst of its previous 3 values.  All models are trained for up to 50 epochs with early stopping. For regularization, dropout with 0.3 probability is applied on the output of all LSTM layers~\cite{Pham2014}.

For inference, we use a greedy decoder to generate the linearized parse\klrm{ sequence}. The output token with maximum posterior probability is chosen at every time step and fed as input in the next time step. The decoder stops upon producing the end-of-sentence symbol. We use TensorFlow~\cite{tensorflow} to implement all models.\footnote{Our code resources can be found in Appendix \ref{app:misc}.}

%% file: result.tex
\section{Results}
\label{sec:results}
\subsection{Text-only Results}
\label{ssec:results-text}

\begin{table}[hbpt]
\centering
\begin{tabular}{l|c|c|c|c}
\hline
Model & F1 & flat-F1 & fluent & disf \\ 
\hline \hline
Berkeley & 85.41 & 85.91 & 90.52 & 83.08 \\ 
C-attn & 83.33 & 83.20 & 90.86 & 79.94 \\
\textbf{CL-attn} & \textbf{87.85} & \textbf{87.68} & \textbf{92.07} & \textbf{85.95} \\\hline
\end{tabular}
\caption{Scores of text-only models on the dev set: 2044 fluent and 3725 disfluent sentences. C-attn denotes \emph{content}-only attention; CL-attn denotes \emph{content+location} attention.}
\label{tab:text_attn}
\end{table}

We first show our results on the model using only text (i.e. $\vec{x}_i=\vec{e}_i$) to establish a strong baseline, on top of which we can add acoustic-prosodic features. 
We experiment with the \emph{content}-only attention model used by~\newcite{Vinyals2015} and the \emph{content+location} attention \klrm{model proposed by}of~\newcite{Chorowski2015}. For comparison with previous non-neural models, we use a high-quality latent-variable parser, the Berkeley parser \cite{Petrov2006}, retrained on our Switchboard data. Table~\ref{tab:text_attn} compares the three text-only models. In terms of F1, the \emph{content+location} attention beats the Berkeley parser by about 2.5\% and \emph{content}-only attention by about 4.5\%. Flat-F1 scores for both encoder-decoder models is lower than their corresponding F1 scores, suggesting that the encoder-decoder models do well on predicting the internal structure of EDIT nodes while the reverse is true for the Berkeley parser.

To explain the gains of \emph{content+location} attention over \emph{content}-only attention, we compare their scores on fluent (without EDIT nodes) and disfluent sentences, shown in Table~\ref{tab:text_attn}. It is clear that most of the gains for \emph{content+location} attention are from disfluent sentences. A possible explanation is the presence of duplicate words or phrases in disfluent sentences, which can be problematic for a \emph{content}-only attention model. Since our best model is the \emph{content+location} attention model, we will henceforth refer to it as the ``CL-attn'' text-only model. 
All models using acoustic-prosodic features are extensions of this model, which provides a strong text-only baseline.

\begin{table}[t]
\centering
\begin{tabular}{l|c|c}
\hline
Model  & Parse & Disf \\ \hline \hline
Berkeley (text only) &  85.41  &  62.45 \\ 
CL-attn (text only) & 87.85 & 79.50\\\hline
CL-attn text and & \\
\ + $p$ & 88.37 &  80.24 \\
\ + $\delta$ & 88.04 & 77.41 \\
\ + $p$ + $\delta$ & 88.21 & 80.84  \\ 
\ + f0/E-CNN & 88.52 & 80.81  \\
\ + $p$ + f0/E-CNN & 88.45 & \textbf{81.19} \\ 
\ + $\delta$ + f0/E-CNN & 88.44 & 80.09 \\
\textbf{ + $\vec{p}$ + $\vec{\delta}$ + f0/E-CNN} & \textbf{88.59} & 80.84 \\
\hline         
\end{tabular}
\centering
\caption{Parse and disfluency detection F1 scores on the dev set. Flat-F1 scores were consistently 0.1\%-0.3\% lower for our models, but 0.2\% higher for the Berkeley parser (85.64).
\label{tab:results-dev}}
\end{table}
\subsection{Adding Acoustic-Prosodic Features}
We extend our CL-attn model with the three kinds of acoustic-prosodic features: pause ($p$), word duration  ($\delta$), and  CNN mappings of fundamental frequency (f0) and energy (E) features (f0/E-CNN).

The results of several model configurations on our dev set are presented in Table~\ref{tab:results-dev}.
First, we note that adding any combination of acoustic-prosodic features (individually or in sets) improves performance over the text-only baseline. 
However, certain combinations of acoustic-prosodic features are not always better than their subsets.
The \emph{text + $p$ + $\delta$ + f0/E-CNN} model that uses all three types of features has the best performance with a gain of 0.7\% over the  already-strong text-only baseline. We will henceforth refer to the \emph{text + $p$ + $\delta$ + f0/E-CNN} model as our ``best model''. 

As a robustness check, we report results of averaging 10 runs on the CL-attn text-only and the best model in Table \ref{tab:variance}. We performed a bootstrap test \cite{efron93} that simulates $10^{5}$ random test draws on the models giving median performance on the dev set. These \emph{median} models gave a statistically significant difference between the text-only and best model ($p$-value $<0.02$). Additionally, a simple t-test over the two sets of 10 results also shows statistical significance $p$-value $<0.03$. 

\begin{table}[t]
\centering
\begin{tabular}{l|c|c}
\hline
Model & Parse & Disf \\ \hline \hline
CL-attn & 87.79 (0.11) & 78.65 (0.46)\\\hline
best model & 88.15 (0.41) & 80.48 (0.70) \\\hline
\end{tabular}
\centering
\caption{Parse and disfluency detection F1 scores on the dev set: mean (and standard deviation) over 10 runs for the baseline text-only model (CL-attn) and the best model with prosody.
}
\label{tab:variance}
\end{table}

\begin{table}[t]
\centering
\begin{tabular}{l|c|c}
\hline
Model  & Parse & Disfl \\ \hline \hline
Berkeley  &  85.87  &  63.44 \\ 
CL-attn & 87.99 &  76.69 \\
{\bf best model} & {\bf 88.50} & {\bf 77.47} \\\hline 
\end{tabular}
\centering
\caption{Parse and disfluency detection F1 scores on the test set. The best model has 
\emph{statistically significant} gains over the text-only baseline with $p$-value $< 0.02$.}
\label{tab:results-test1}
\end{table}

Table~\ref{tab:results-test1} presents the results on the test set. Again, adding the acoustic-prosodic features improves over the text-only baseline. 
The gains are statistically significant for the best model with $p$-value $< 0.02$, again using a bootstrap test with simulated $10^{5}$ random test draws on the two models. 

Table~\ref{tab:results-test2} includes results from prior studies that compare systems using text alone with ones that incorporate prosody, given hand transcripts and sentence segmentation. It is difficult to compare systems directly, because of the many differences in the experimental set-up. For example, the original 
\newcite{Charniak2001} 
result (reporting F=85.9 for parsing and F=78.2 for disfluencies) leverages punctuation in the text stream, which is not realistic for speech transcripts and not used in most other work. Our work benefits from more text training material than others, but others benefit from gold part-of-speech tags. 
\newcite{Kahn2005} 
use a modified sentence segmentation. There are probably minor differences in handling of word fragments and scoring edit regions. Thus, this table primarily shows that our framework leads to more benefits from sentence-internal prosodic cues than others have obtained.

\begin{table}[t]
\centering
\begin{tabular}{l|c|c}
\hline
Model  & Parse & Disfl \\ \hline \hline
Text Only & & \\
\hspace{0.1in} Kahn et al. \shortcite{Kahn2005} & 86.4~~ & 78.2 \\
\hspace{0.1in} Hale et al. \shortcite{Hale2006} & 71.16 & 41.7 \\
\hspace{0.1in} CL-attn (text only) & 87.99 &  76.7 \\
\hline
Text + Prosody & & \\
\hspace{0.1in} Kahn et al. \shortcite{Kahn2005} & 86.6~~ & 78.2 \\
\hspace{0.1in} Hale et al. \shortcite{Hale2006} & 71.05 & 36.2 \\
\hspace{0.1in} best model & 88.50  & 77.5 \\\hline         
\end{tabular}
\centering
\caption{Parse and disfluency detection F1 scores on the test set comparing to other reported results. 
}
\label{tab:results-test2}
\end{table}

%% file: analysis.tex
\section{Analysis}
\label{sec:analysis}
\begin{figure}[t]
\centering

\includegraphics[
width=0.45\textwidth]{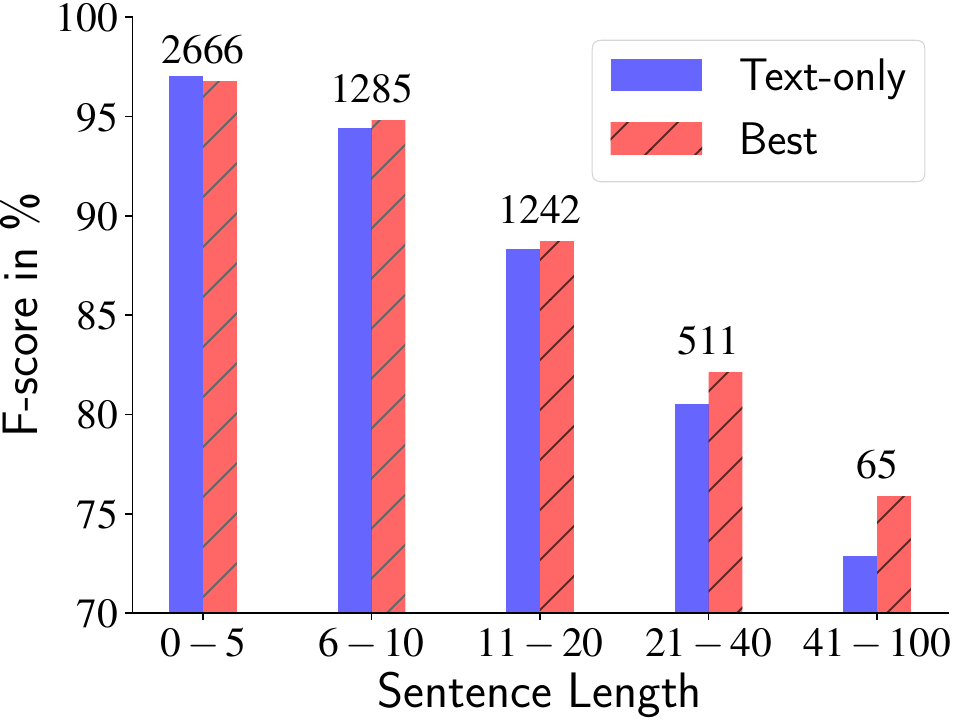}
\vspace{-0.05in}
\caption{F1 scores of the text-only model and our best model as a function of sentence length. 
\label{fig:error}
}
\end{figure}

\begin{figure*}[h]
    \includegraphics[width=\textwidth]{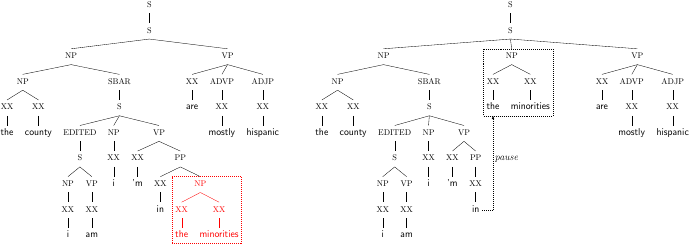}
    \caption {An example sentence from development data -- \textit{the county i am i 'm in [pause] the minorities are mostly hispanic}.  The text-only parser (on the left)\klrm{, which is oblivious to prosodic cues,} makes a PP Attachment error.  The prosody-enhanced parser (on the right) uses the pause indicator to correctly predict a constituent change after the word \emph{in}.}
    \label{fig:tree}
\end{figure*}

\paragraph{Effect of sentence length.} Figure~\ref{fig:error} 
shows performance differences between our best model and the text-only model for varying sentence lengths. Both models do worse on longer sentences, as expected since the corresponding parse trees tend to be more complex.  The performance difference between our best model and the text-only model increases with sentence length. This is likely because longer sentences more often have multiple prosodic phrases and disfluencies.

\begin{table}[t]
\centering
\begin{tabular}{l|c|c}
\hline
Model  & fluent & disfluent \\ \hline \hline
text-only &  92.07 & 85.90 \\
best model & 92.03 &  \textbf{87.02}\\\hline
\end{tabular}
\caption{Dev set F1-score of text-only and best model on fluent (2029) vs.\ disfluent (3689) sentences.\textsuperscript{\ref{subset}}}
\label{tab:text_vs_best}
\end{table}

\paragraph{Effect of disfluencies.} Table \ref{tab:text_vs_best} presents parse scores on the subsets of fluent  and disfluent sentences, showing that the performance gain is in the disfluent set (65\% of the dev set sentences). Because sentence boundaries are given, and so many fluent sentences in spontaneous speech are short, there is less potential for benefit from prosody in the fluent set.

\begin{table}[t]
\centering
\begin{tabular}{l|r|r}
\hline 
\multirow{2}{*}{Error Type} & \multicolumn{2}{c}{Disfluent Sentences} \\
 & text + $p$ & best model \\ \hline \hline
Clause Att. & 5.7\% & 1.3\% \\
Diff. Label & 7.6\% & 4.2\% \\
Modifier Att. & 9.7\% & 19.1\% \\
NP Att. & -2.7\% & 14.5\% \\
NP Internal & 7.8\% & 7.4\% \\
PP Att. & 10.1\% & 7.8\% \\
1-Word Phrase & 6.3\% & 6.8\% \\
Unary & -1.1\% & 8.9\% \\
VP Att. & 0.0\% & 12.0\% \\\hline
\end{tabular}
\caption{Relative error reduction over the text-only baseline in the disfluent subset (3689 sentences) of the development set. Shown here are the most frequent error types (with count $\ge 100$ for the text-only model).} 
\label{tab:error_rel}
\end{table}

\paragraph{Types of errors.} We use the Berkeley Parser Analyzer \cite{Kummerfeld2012} to compare the types of errors made by the different parsers.\footnote{This analysis omits the 1\% of the sentences that did not have timing information.\label{subset}}
Table~\ref{tab:error_rel} presents the relative error reductions over the text-only baseline achieved by the text + $p$ model and our best model for disfluent sentences. The two models differ in the types of error reductions they provide.
Including pause information gives largest improvements on PP attachment and Modifier attachment errors. Adding the remaining acoustic-prosodic features helps to correct more types of attachment errors, especially VP and NP attachment.
Figure~\ref{fig:tree} demonstrates one case where the pause feature helps in correcting a PP attachment error made by a text-only parser. Other interesting examples (see Appendix \ref{app:examples}) suggest that the learned f0/E features help reduce NP attachment errors where the audio reveals a prominent word at the constituent boundary, even though there is no pause at that word.

\paragraph{Effect of transcription errors.}  The results and analyses so far have assumed that we have reliable transcripts. In fact, the original transcripts contained errors, and the Treebank annotators used these without reference to audio files. Mississippi State University (MS-State) ran a clean-up project that produced more accurate word transcripts and time alignments \cite{Deshmukh1998}.
The NXT corpus provides reconciliation between Treebank and MS-State transcripts in terms of annotating missed/extra/substituted words, but parses were not re-annotated. The transcript errors mean that the acoustic signal is inconsistent with the ``gold'' parse tree. Below are some examples of ``fluent'' sentences (according to the Treebank transcripts) with transcription errors, for which prosodic features ``hurt'' parsing. Words that transcribers missed are in brackets and those inserted are underlined.
\begin{description}
    \item[S1:] \sffamily{\small and because $<$uh$>$ like if your spouse died $<$all of a sudden you be$>$ all alone it 'd be nice to go someplace \underline{with} people similar to you to have friends}
    \item[S2:] \sffamily{\small uh \underline{uh} $<$i have had$>$ my wife 's picked up a couple \underline{of} things saying uh boy if we could refinish that 'd be a beautiful piece of furniture}
\end{description}
Multi-syllable errors are especially problematic, leading to serious inconsistencies between the text and the acoustic signal. Further, the missed words lead to an incorrect attachment in the ``gold'' parse in S1 and a missing restart edit in S2.
Indeed, for sentences with consecutive transcript errors, which we expect to impact the prosodic features, there is a statistically significant ($p$-value $<0.05$) negative effect on parsing with prosody. 
Not included in this analysis are sentence boundary errors, which also change the ``gold'' parse. Thus, prosody may be more useful than results here indicate.

%% file: related.tex
\section{Related Work}
\label{sec:related}

Related work on parsing conversational speech has mainly addressed four problems: speech recognition errors, unknown sentence segmentation, disfluencies, and integrating prosodic cues. Our work addresses the last two problems, which involve studies based on hand-transcribed text and known sentence boundaries, as in much speech parsing work. The related studies are thus the focus of this discussion. We describe studies using the Switchboard corpus, since it has dominated work in this area, being the largest source of treebanked English spontaneous speech.

One major challenge of parsing conversational speech is the presence of disfluencies, which are grammatical and prosodic interruptions. Disfluencies include repetitions (`I am + I am'), repairs (`I am + we are'), and restarts (`What I + Today is the...'), where the `+' corresponds to an interruption point. Repairs often involve parallel grammatical constructions, but they can be more complex, involving hedging, clarifications, etc. Charniak and Johnson \cite{Charniak2001,Johnson2004} demonstrated that disfluencies are different in character than other constituents and that parsing performance improves from combining a PCFG parser with a separate module for disfluency detection via parse rescoring. Our approach does not use a separate disfluency detection module; we hypothesized that the location-sensitive attention model helps handle these differences based on analysis of the text-only results (Table~\ref{tab:text_attn}). 
However, more explicit modeling of disfluency pattern match characteristics in a dependency parser \cite{Honnibal2014} leads to better disfluency detection performance (F = 84.1 vs.\ 76.7 for our text only model). Pattern match features also benefit a neural model for disfluency detection alone (F~=~87.0) \cite{Zayats2016}, 
and similar gains are observed by formulating disfluency detection in a transition-based framework (F~=~87.5) \cite{Wang2017}.
Experiments with oracle disfluencies as features improve the CL-attn text-only parsing performance from 87.85 to 89.38 on the test set, showing that more accurate disfluency modeling is a potential area of improvement.

It is well known that prosodic features play a role in human resolution of syntactic ambiguities, with more than two decades of studies seeking to incorporate prosodic features in parsing. A series of studies looked at constituent parsing informed by the presence (or likelihood) of prosodic breaks at word boundaries \cite{Kahn2004,Kahn2005,Hale2006,Dreyer2007}. Our approach improves over performance of these systems using raw acoustic features, without the need for hand-labeling prosodic breaks. The gain is in part due to the improved text-based parser, but the incremental benefit of prosody here is similar to that in these prior studies. (In prior work using acoustic feature directly \cite{Gregory2004}, prosody actually degraded performance.) 
Our analyses of the impact of prosody also extends prior work.

Prosody is also known to provide useful cues to sentence boundaries \cite{liu06}, and   automatic sentence segmentation performance has been shown to have a significant impact on parsing performance \cite{Kahn2012}. 
In our study, sentence boundaries are given so as to focus on the role of prosody in resolving sentence-internal parse ambiguity, for which prior work had obtained smaller gains.
Studies have also shown that parsing lattices or confusion networks can improve ASR performance \cite{Kahn2012,Yoshikawa2016}. Our analysis of performance degradation for the system with prosody when the gold transcript and associated parse are in error suggests that prosody may have benefits for parsers operating on alternative ASR hypotheses.

The results we compare to in Section~\ref{sec:results} are relatively old. More recent parsing results on spontaneous speech involve dependency parsers using only text \cite{Rasooli2013,Honnibal2014,Yoshikawa2016}, with the exception of a study on unsupervised dependency parsing \cite{Pate2013}. With the recent success of transition-based neural approaches in dependency parsing, researchers have adapted transition-based ideas to constituent parsing \cite{Zhu2013,Watanabe2015,Dyer2016}. These approaches have not yet been used with speech, to our knowledge, but we expect it to be straightforward to extend our prosody integration framework to these systems, both for dependency and constituency parsing.

%% file: conclusion.tex
\section{Conclusion}
\label{sec:conclusion}
We have presented a framework for directly integrating acoustic-prosodic features with text in a neural encoder-decoder parser that does not require hand-annotated prosodic structure. On conversational sentences, we obtained strong results when including word-level acoustic-prosodic features over using only transcriptions. The acoustic-prosodic features provide the largest gains when sentences are disfluent or long, and analysis of error types shows that these features are especially helpful in repairing attachment errors. In cases where prosodic features hurt performance, we observe a statistically significant negative effect caused by imperfect human transcriptions
that make the ``ground truth'' parse tree and the acoustic signal inconsistent, which suggests that there is more to be gained from prosody than observed in prior studies. We thus plan to investigate aligning the Treebank and MS-State versions of Switchboard for future work.

Here, we assumed known sentence boundaries and hand transcripts, leaving open the question of whether increased benefits from prosody can be gained by incorporating sentence segmentation in parsing and/or in parsing ASR lattices.
Most prior work using prosody in parsing has been on constituent parsing, since prosodic cues tend to align with constituent boundaries.
However, it remains an open question as to whether dependency, constituency or other parsing frameworks are better suited to leveraging prosody.    
Our study builds on a parser that uses reverse order text processing, since it provides a stronger text-only baseline. However, the prosody modeling component relies only on a 1 second lookahead of the current word (for pause binning), so it could be easily incorporated in an incremental parser.

%% file: acknowledgement.tex
\section*{Acknowledgement}
We thank the anonymous reviewers for their helpful feedback. We also thank Pranava Swaroop Madhyastha, Hao Tang, Jon Cai, Hao Cheng, and Navdeep Jaitly for their help with initial discussions and code setup. This research was partially funded by a Google Faculty Research Award to Mohit Bansal, Karen Livescu, and Kevin Gimpel; and NSF grant no.\ IIS-1617176. The opinions expressed in this work are those of the authors and do not necessarily reflect the views of the funding agency.

%% file: appendix.tex
\section{Appendix}
\label{sec:appendix}

\input{misc}
\input{tree-examples}

%% file: misc.tex
\subsection{Miscellany}
\label{app:misc}
\ttcomment{Added code repo links}
Our main model code is available at \url{https://github.com/shtoshni92/speech_parsing}. Most of the data preprocessing code is available at \url{https://github.com/trangham283/seq2seq_parser/tree/master/src/data_preps}. Part of our data preprocessing pipeline also uses  \url{https://github.com/syllog1sm/swbd_tools}.

Table \ref{tab:data_stats} shows statistics of our Switchboard dataset. As defined, for example, in \cite{Charniak2001,Honnibal2014}, the splits are: conversations sw2000 to sw3000 for training, sw4500 to sw4936 for validation (dev), and sw4000 to sw4153 for evaluation (test). In addition, previous work has reserved sw4154 to sw4500 for ``future use'' (dev2), but we added this set to our training set. That is, all of our models are trained on Switchboard conversations sw2000 to sw3000 as well as sw4154 to sw4500. 

\begin{table}[hbpt]
\centering
\begin{tabular}{l|r|r}
\hline
Section  & \# sentences & \# words\\ \hline \hline
Train  &  97,113  &  729,252 \\ 
Dev & 5,769 &  50,445 \\
Test & 5,901  & 48,625 \\\hline         
\end{tabular}
\centering
\caption{Data statistics.}
\label{tab:data_stats}
\end{table}

\begin{figure}[hbpt]
\centering
\includegraphics[width=0.48\textwidth]{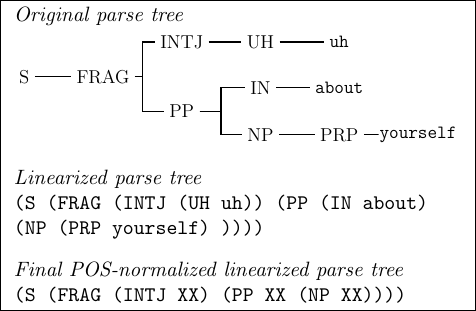}
\centering
\caption{Data preprocessing. Trees are linearized; POS tags (pre-terminals) are normalized as ``XX''. Also note the annotation standard used for Switchboard data: The root node of the tree is an ``S'' node although it is not a complete sentence.}
\label{fig:example}
\end{figure}

Figure \ref{fig:example} illustrates the data preprocessing step. On the decoder end, we also use a post-processing step that merges the original sentence \klrm{tokens }with the decoder output to obtain the standard constituent tree representation. During inference, in rare cases (and virtually none as our models converge), the decoder does not generate a valid parse sequence, due to the mismatch in brackets and/or the mismatch in the number of pre-terminals and terminals, i.e., num(XX) $\ne$ num(tokens). In such cases, we simply add/remove brackets from either end of the parse, or add/remove pre-terminal symbols XX in the middle of the parse to match the number of input tokens. 

\begin{figure}[hbpt]
\centering
\includegraphics[width=0.48\textwidth]{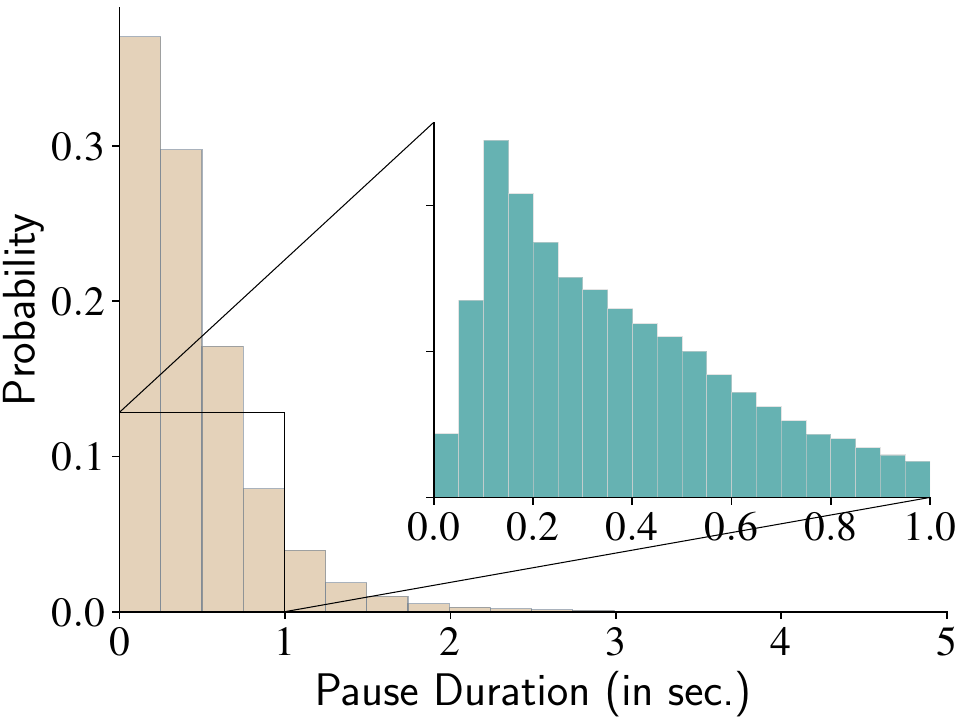}
\caption{Histogram of inter-word pause durations in our training set. As expected, most of the pauses are less than 1 second. Further binning of pause durations $\le$ 1 second reveals that the plot peaks around 0.2 seconds and continuously decays from there on. In some very rare cases, pauses of 5+ seconds occur within a sentence.} 
\label{fig:pause_dur}
\end{figure}

Figure~\ref{fig:pause_dur} shows the distribution of pause durations in our training data. Our pause buckets of $0 < p \le 0.05$ s, $0.05 \; \textrm{s} < p \le 0.2$ s, $0.2 < p \le 1$ s, and $p > 1$ s described in the main paper were based on this distribution of pause lengths.

Table \ref{tab:parse_err} shows the comprehensive error counts in all error categories defined in the Berkeley Parse Analyzer \cite{Kummerfeld2012} in both the fluent and disfluent subsets.

\begin{table*}[thbp]
\centering
\begin{tabular}{l||r|r|r||r|r|r}
\hline
 & \multicolumn{3}{c||}{Fluent Subset} & \multicolumn{3}{c}{Disfluent Subset} \\ \cline{2-7}
Error Type & text-only & text + $p$ & best model & text-only & text + $p$ & best model \\ \hline \hline 
Clause Attach. & 126 & 132 & 123 & 631 & 595 & 600 \\
Co-ordination & 1 & 2 & 1 & 10 & 10 & 5 \\
Different label & 112 & 116 & 124 & 288 & 266 & 300 \\
Modifier Attach. & 119 & 127 & 112 & 320 & 289 & 325 \\
NP Attach. & 92 & 89 & 94 & 332 & 341 & 283 \\
NP Internal & 71 & 61 & 65 & 231 & 213 & 232 \\
PP Attach. & 171 & 152 & 149 & 524 & 471 & 470 \\
1-Word Phrase & 334 & 342 & 328 & 1054 & 988 & 1030 \\
UNSET add & 86 & 81 & 64 & 353 & 352 & 356 \\
UNSET move & 85 & 93 & 95 & 466 & 447 & 439 \\
UNSET remove & 73 & 70 & 56 & 334 & 324 & 318 \\
Unary & 246 & 239 & 236 & 1088 & 1100 & 1074 \\
VP Attach. & 36 & 41 & 25 & 167 & 167 & 172 \\
XoverX Unary & 36 & 35 & 34 & 54 & 57 & 54 \\
\hline
\end{tabular}
\caption{Parse error counts comparison on the fluent (2029 sentences) and disfluent (3689 sentences) subsets of the development set across three parsers.}
\label{tab:parse_err}
\end{table*}

%% file: tree-examples.tex
\subsection{Tree Examples}
\label{app:examples}

\begin{figure*}[hbpt]
\centering
\includegraphics[trim={1cm 11cm 1cm 0},clip,scale=0.5]{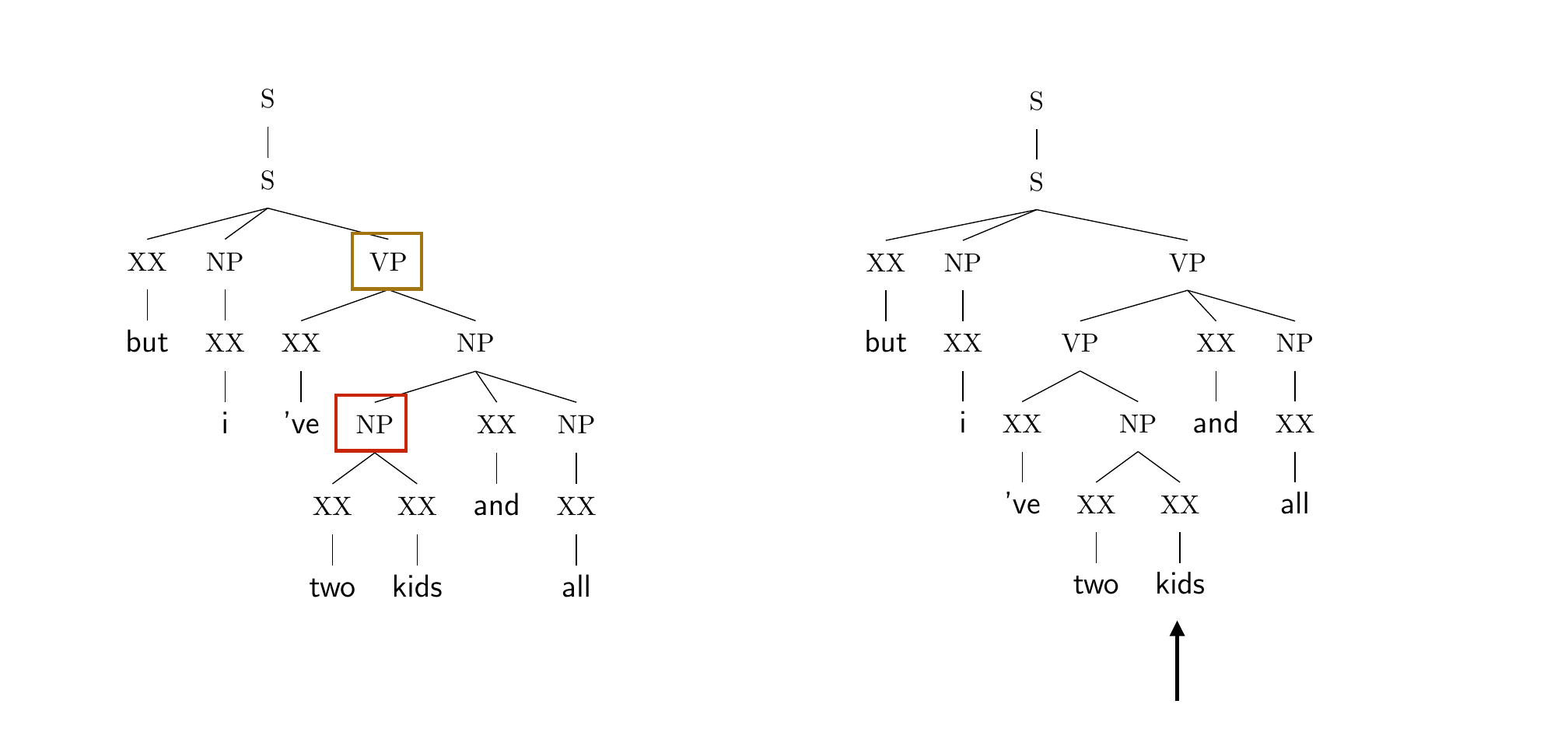}
\centering
\caption{An example sentence from development data -- \textit{but i 've two \underline{kids} and all}. Even though there are no pauses between all words, the word \emph{kids} is lengthened in the audio sample, helping the prosody-enhanced parser (right) to recognize a major syntactic boundary, avoiding the NP Attachment error made by the text-only parser (left).}
\label{fig:app_1}
\end{figure*}

\begin{figure*}[hbpt]
\centering
\includegraphics[trim={1cm 9cm 1cm 0},clip,scale=0.5]{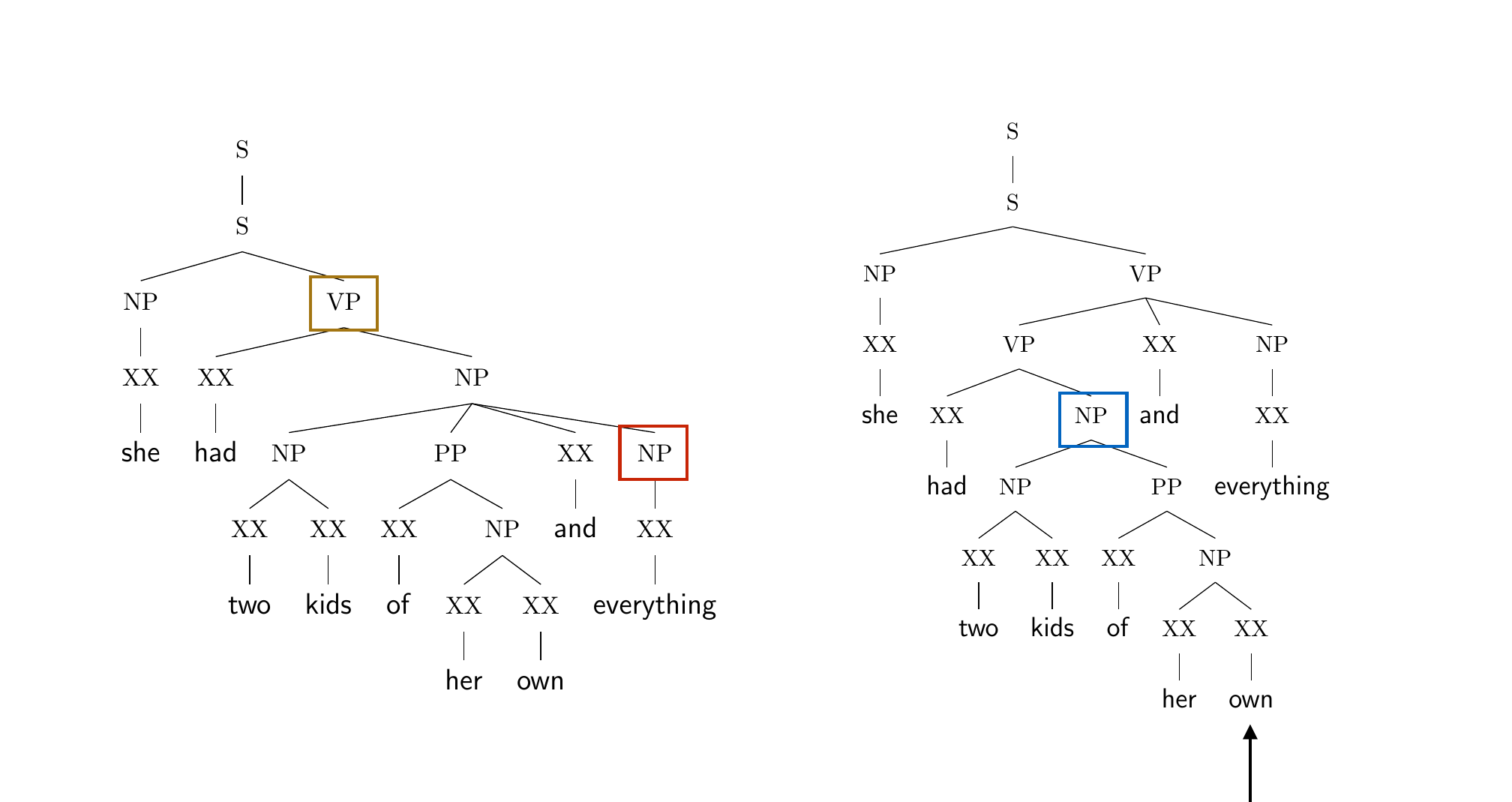}
\centering
\caption{An example sentence from development data -- \textit{she had two kids of her \underline{own} and everything}. There were no pauses between all words in this sentence, the audio sample showed that the word \emph{own} was both lengthened and raised in intonation, giving the prosody-enhanced parser (right) a signal that \emph{own} is on a syntactic boundary. On the other hand, the text-only parser (left) had no such information and made an NP-attachment error. This sentence also illustrates an interesting case where, in isolation, the text-only parse makes sense (i.e. \emph{everything} being an object of \emph{had}). However, in the context of this conversation (the speaker was talking about another person in an informal manner), \emph{and everything} acts more like filler - e.g. ``i play the violin \emph{and stuff}''}
\label{fig:app_2}
\end{figure*}

\begin{figure*}[hbpt]
\centering
\includegraphics[trim={1cm 10cm 1cm 1cm},clip,scale=0.46]{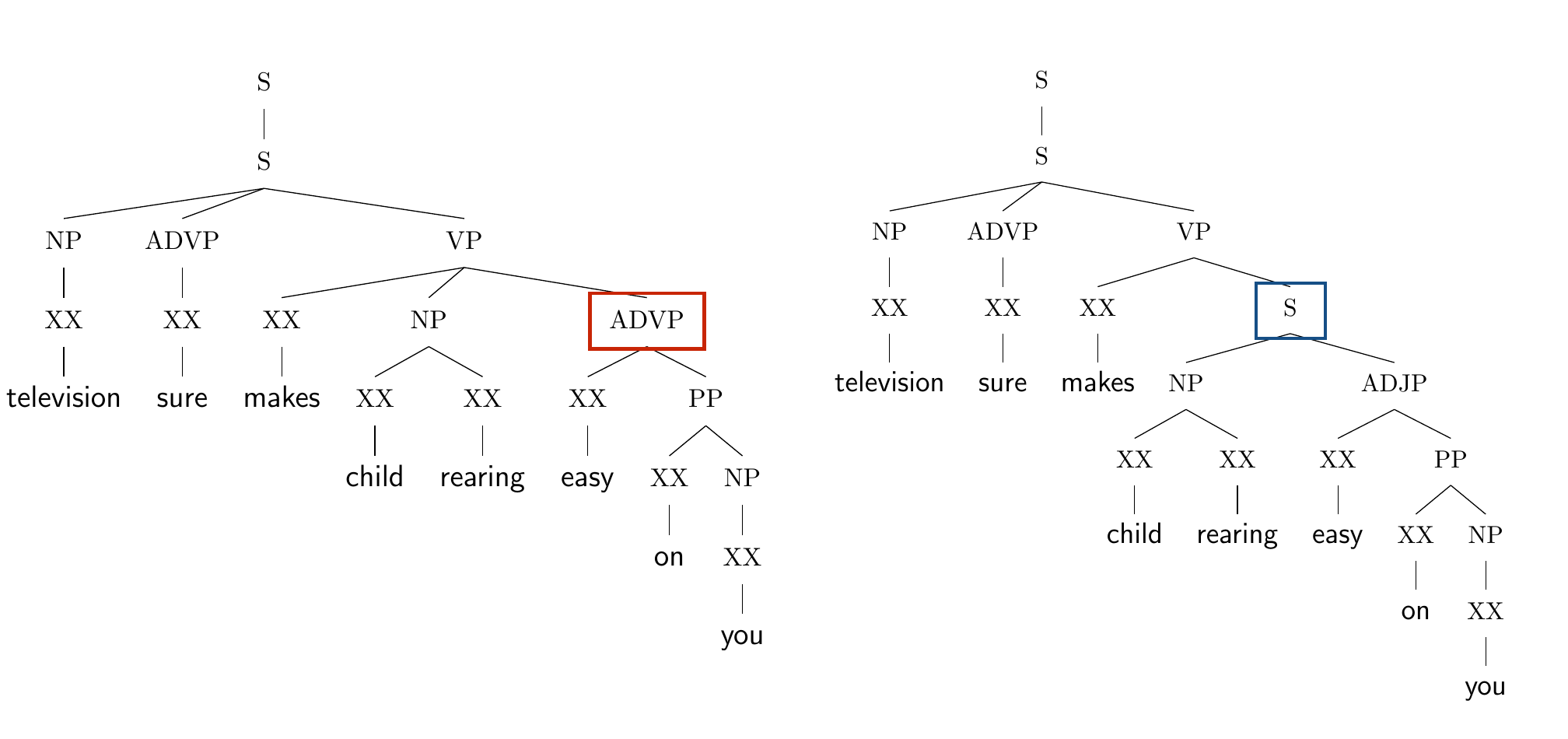}
\centering
\caption{An example sentence from development data -- \textit{television sure makes child rearing easy on you}. This is an example where our prosody-enhanced parser (left) did worse than the text-only parser (right), which made no errors. The error type illustrated here is Different Label and Modifier Attachment. In the first iteration, the analyzer identifies a Different Label error (ADVP node), and in the second pass identifies the Modifier Attachment error.}
\label{fig:app_3}
\end{figure*}

In figures~\ref{fig:app_1},~\ref{fig:app_2}, and ~\ref{fig:app_3}, we follow node correction notations as in ~\cite{Kummerfeld2012}. 
In particular, missing nodes are marked in blue on the gold tree, extra nodes are marked red in the predicted tree, and yellow nodes denote crossing. 

